\documentclass{article}
\usepackage{ijcai20}

\usepackage{times}
\usepackage{soul}
\usepackage{url}
\usepackage[hidelinks]{hyperref}
\usepackage[utf8]{inputenc}
\usepackage[small]{caption}
\usepackage{graphicx}
\usepackage{subfigure}
\usepackage{paralist,amsmath, amssymb,amsthm,bbm,mathtools}
\usepackage{booktabs}
\usepackage{lineno}
\usepackage{algorithm,algorithmicx}
\usepackage{algpseudocode}
\usepackage{multirow}
\usepackage{bm}
\usepackage[normalem]{ulem}
\useunder{\uline}{\ul}{}

\newtheorem{thm}{Theorem}
\newtheorem{prop}{Proposition}
\newtheorem{lem}{Lemma}

\newtheorem{ass}{Assumption}
\newtheorem*{rem}{Remark}
\newtheorem{exam}{Example}

\title{Nearly Optimal Regret for Stochastic Linear Bandits with Heavy-Tailed Payoffs}

\author{
   Bo Xue\and 
   Guanghui Wang\and 
   Yimu Wang \And 
   Lijun Zhang \footnote{Lijun Zhang is the corresponding author.}
   \affiliations
   National Key Laboratory for Novel Software Technology, Nanjing University, Nanjing 210023, China
   \emails
   \{xueb, wanggh, wangym, zhanglj\}@lamda.nju.edu.cn
}

\DeclareMathOperator*{\argmin}{argmin}
\DeclareMathOperator*{\argmax}{argmax}
\DeclareMathOperator*{\ie}{i.e.}
\DeclarePairedDelimiter\norm{\lVert}{\rVert}

\def \E {\mathrm{E}}
\def \N {\mathrm{N}}
\def \R {\mathbb{R}}

\def \P {\mathrm{P}}
\def \Q {\mathrm{Q}}
\def \B {\mathcal{B}}
\def \A {\mathcal{A}}

\def \r {\mathbf{r}}
\def \ones {\mathbf{1}}

\def \Var {\mathrm{Var}}

\begin{document}

\maketitle

\begin{abstract} 
In this paper, we study the problem of stochastic linear bandits with finite action sets. Most of existing work assume the payoffs are bounded or sub-Gaussian, which may be violated in some scenarios such as financial markets. To settle this issue, we analyze the linear bandits with heavy-tailed payoffs, where the payoffs admit finite $1+\epsilon$ moments for some $\epsilon\in(0,1]$. Through median of means and dynamic truncation, we propose two novel algorithms which enjoy a sublinear regret bound of $\widetilde{O}(d^{\frac{1}{2}}T^{\frac{1}{1+\epsilon}})$, where $d$ is the dimension of contextual information and $T$ is the time horizon. Meanwhile, we provide an $\Omega(d^{\frac{\epsilon}{1+\epsilon}}T^{\frac{1}{1+\epsilon}})$ lower bound, which implies our upper bound matches the lower bound up to polylogarithmic factors in the order of $d$ and $T$ when $\epsilon=1$. Finally, we conduct numerical experiments to demonstrate the effectiveness of our algorithms and the empirical results strongly support our theoretical guarantees.
\end{abstract}

\section{Introduction}\label{intro}
Bandit online learning is a powerful framework for modeling various important decision-making scenarios with applications ranging from medical trials to advertisement placement to network routing \cite{Bandit:suvery}. In the basic stochastic multi-arm bandits (MAB) \cite{Robbins:52}, a learner repeatedly selects one from $K$ arms to play, and then observes a payoff drawn from a fixed but unknown distribution associated with the chosen arm. The learner's goal is to maxmize the cumulative payoffs through the trade-off between exploration and exploitation, i.e., pulling the arms that may potentially give better outcomes and playing the optimal arm in the past \cite{Auer:2002}. The classic upper confidence bound (UCB) algorithm achieves a regret bound of $O(K\log T)$ over $T$ iterations and $K$ arms, which matches the minimax regret up to a logarithmic factor \cite{Lai1985}.

One fundamental limitation of stochastic MAB is that it ignores the side information (contexts) inherent in the aforementioned real-world applications, such as the user and webpage features in advertisement placement \cite{abe2003reinforcement}, which could guide the decision-making process. To address this issue, various algorithms have been developed to exploit the contexts, based on different structures of the payoff functions such as Lipschitz \cite{Kleinberg:2008:MBM,Bubeck:2011} or convex \cite{Agarwal:2013,Bubeck:2015}. Among them, the stochastic linear bandits (SLB) has received significant research interests \cite{Auer:2002,Chu:2011}, in which the expected payoff at each round is assumed to be a linear combination of features in the context vector. More precisely, in each round of SLB, the learner first observes feature vector $x_{t,a}\in\R^d$ for each arm $a$. After that, he/she selects an arm $a_t$ and receives payoff $r_{t,a_t}$, such that
\begin{equation}\label{expected_payoff}
\E[r_{t,a_t}|x_{t,a_t}]=x_{t,a_t}^\top\theta_*
\end{equation}
where $\theta_*\in\R^d$ is a vector of unknown parameters. The metric to measure the learner's performance is expected regret, defined as
\begin{equation*}\label{regret}
R(T)= \sum_{t=1}^T  x_{t,a_t^*}^\top\theta_*- \sum_{t=1}^ T x_{t,a_t}^\top\theta_*
\end{equation*}
where $a_t^*=\argmax_{a\in\{1,2,\ldots,K\}}x_{t,a}^\top\theta_*$ and $a_t$ is the action chosen by the learner at round $t$.

While SLB has been explored extensively \cite{Auer:2002,Chu:2011,Abbasi:2011,Zhang:2016}, most of the previous work assume the payoffs are bounded or satisfy the sub-Gaussian property. However, in many real-world scenarios such as financial markets \cite{Cont:2000} and  neural oscillations \cite{Roberts:2015}, the payoffs $r_{t,a}$ fluctuate rapidly and do not exhibit bounded or sub-Gaussian property but satisfy heavy-tailed distributions  \cite{Foss:2013}, $\ie$,
$$\lim_{c \to \infty} \mathbb{P}\left\{r_{t,a}-\E[r_{t,a}]>c\right\}\cdot e^{\lambda c}=\infty, \ \ \forall \lambda>0.$$
There exists a rich body of work on learning with heavy-tailed distribution \cite{Audibert:2011,Catoni:2012,Brownlees:2015,Hsu:2016,Zhang:2018,Lu:2019}, but limited work contributed to the setting of stochastic linear bandits. \citeauthor{Medina:2016} \shortcite{Medina:2016} is the first to investigate this problem, and develop two algorithms achieving  $\widetilde{O}(dT^{\frac{2+\epsilon}{2(1+\epsilon)}})$ and $\widetilde{O}(\sqrt{d}T^{\frac{1+2\epsilon}{1+3\epsilon}}+dT^{\frac{1+\epsilon}{1+3\epsilon}})$ regret bounds respectively under the assumption that the distributions have finite moments of order $1+\epsilon$ for some $ \epsilon \in (0, 1]$. Later, \citeauthor{Han:2018} \shortcite{Han:2018} improve these bounds to $\widetilde{O}(dT^{\frac{1}{1+\epsilon}})$ by developing two more delicate algorithms. When the variance of payoff is finite ($\ie$, $\epsilon = 1$), this bound becomes $\widetilde{O}(d\sqrt{T})$, which is nearly optimal in terms of $T$. However, when the number of arms is finite, this upper bound is sub-optimal as there exists an $O(\sqrt{d})$ gap from the lower bound $\Omega(\sqrt{dT})$ \cite{Chu:2011}. Thus, an interesting challenge is to recover the regret of $O(\sqrt{dT})$ under the heavy-tailed setting for linear bandits with finite arms.

To the best of our knowledge, this is the first work which investigates heavy-tailed SLB with finite arms and our contributions are highlighted as follows:
\begin{itemize}
  \item We propose two novel algorithms to address the heavy-tailed issue in stochastic linear bandits with finite arms. One is developed based on median of means, and the other adopts the truncation technique. Furthermore, we establish an $\widetilde{O}(d^{\frac{1}{2}}T^{\frac{1}{1+\epsilon}})$ regret bound for both algorithms.
  \item We provide an $\Omega(d^{\frac{\epsilon}{1+\epsilon}}T^{\frac{1}{1+\epsilon}})$ lower bound for heavy-tailed SLB problem, which matches our upper bound in terms of the dependence on $T$. It also implies the dependence on $d$ in our upper bound is optimal up to a logarithmic term when $\epsilon=1$. 
  \item We conduct numerical experiments to demonstrate the performance of our algorithms. Through comparisons with existing work, our proposed algorithms exhibit improvements on heavy-tailed bandit problem. 
\end{itemize}

\section{Related Work}\label{sec:1}
In this section, we briefly review the related work on bandit learning. The $p$-norm of vector $x\in\R^d$ is $\norm{x}_p=(|x_1|^p+\ldots+|x_d|^p)^{1/p}$ and the $\ell_2$-norm is denoted as $\norm{\cdot}$.

\subsection{Bandit Learning with Bounded/Sub-Gaussian Payoffs}
The celebrated work of \citeauthor{Lai1985} \shortcite{Lai1985} derived a lower bound of $\Omega(K\log T)$ for stochastic MAB, and proposed an algorithm which achieves the lower bound asymptotically by making use of the upper confidence bound (UCB) policies. \citeauthor{Auer:2002} \shortcite{Auer:2002} studied the problem of stochastic linear bandits, and developed a basic algorithm named LinRel to solve this problem. However, he failed to provide a sublinear regret for LinRel since the analysis of the algorithm requires all observed payoffs so far to be independent random variables, which may be violated. To resolve this problem, he turned LinRel to be a subroutine which assumes independence among the payoffs, and then constructed a master algorithm named SupLinRel to ensure the independence. Theoretical analysis demonstrates that SupLinRel enjoys an $\widetilde{O}(\sqrt{dT})$ regret bound, assuming the number of arms is finite. \citeauthor{Chu:2011} \shortcite{Chu:2011} modified LinRel and SupLinRel slightly to BaseLinUCB and SupLinUCB, which enjoy similar regret bound but less computational cost and easier theoretical analysis. They also provided an $\Omega(\sqrt{dT})$ lower bound for SLB. \citeauthor{Linear:Bandit:08} \shortcite{Linear:Bandit:08} considered the setting where the arm set is infinite, and proposed an algorithm named ConfidenceBall$_2$ which enjoys a regret bound of $\widetilde{O}(d\sqrt{T})$. Later, \citeauthor{Abbasi:2011} \shortcite{Abbasi:2011} provided a new analysis of ConfidenceBall$_2$, and improved the worst case bound by a logarithmic factor.

The main diffculty in bandit problem is the trade-off between exploitation and exploration. Most of the existing work take advantage of UCB to settle this issue and adopt the tool of ridge regression to estimate $\theta_*$ \cite{Auer:2002,Chu:2011}. The least square estimator of \citeauthor{Chu:2011} \shortcite{Chu:2011} is
\begin{equation}\label{LSE}
\hat{\theta}_t=\argmin_{\theta\in\R^d}\norm{V_t\theta-Y_t}^2+\norm{\theta}^2
\end{equation}
where $V_t =[x_{\tau,a_\tau}]_{\tau\in\Psi_t}\in\R^{|\Psi_t|\times d}$ is a matrix of the historical contexts, $Y_t=[r_{\tau,a_\tau}]_{\tau\in\Psi_t}\in\R^{|\Psi_t|\times 1}$ is the historical payoff vector and $\Psi_t\subseteq\{1,2,\ldots,t-1\}$ is a filtered index set. The confidence interval for arm $a$ at round $t$ is
\begin{equation}\label{CI}
\left[x_{t,a}^\top\hat{\theta}_t-w_{t,a},\ x_{t,a}^\top\hat{\theta}_t+w_{t,a}\right]
\end{equation}
where $w_{t,a}=(\alpha_t+1)\sqrt{x_{t,a}^\top A_tx_{t,a}}, A_t= I_d+V_t^\top V_t$ and $\alpha_t=O(\sqrt{\ln(TK)})$. If $w_{t,a}$ is small for all $a\in\{1,2,\ldots,K\}$, which means the estimations for coming payoffs are accurate enough, the arm with highest upper confidence bound is played to execute exploitation. Otherwise, if there exists an arm $a$ with $w_{t,a}$ large enough, arm $a$ is played to explore more information.

\subsection{Bandit Learning with Heavy-tailed Payoffs}
The classic paper of \citeauthor{Bubeck:2013} \shortcite{Bubeck:2013} studied stochastic MAB with heavy-tailed payoffs, and proposed a UCB-type algorithm which enjoys a logarithmic regret bound, under the assumption that the $1+\epsilon$ moment of the payoffs is bounded for some $\epsilon\in(0,1]$. They also constructed a matching lower bound. \citeauthor{Medina:2016} \shortcite{Medina:2016} extended the analysis to SLB, and developed two algorithms enjoying $\widetilde{O}(dT^{\frac{2+\epsilon}{2(1+\epsilon)}})$ and $\widetilde{O}(\sqrt{d}T^{\frac{1+2\epsilon}{1+3\epsilon}}+dT^{\frac{1+\epsilon}{1+3\epsilon}})$ regret bounds respectively. In a subsequent work, \citeauthor{Han:2018} \shortcite{Han:2018} constructed an $\Omega(dT^{\frac{1}{1+\epsilon}})$ lower bound for SLB with heavy-tailed payoffs, assuming the arm set is infinite, and developed algorithms with matching upper bounds in terms of $T$. 

An intuitive explanation for heavy-tailed distribution is that extreme values are presented with high probability. One strategy tackling the heavy-tailed problem is median of means \cite{Hsu:2016}, whose basic idea is to divide all samples drawn from the distribution into several groups, calculate the mean of each group and take the median of these means. Another strategy is truncation following the line of research stemmed from \citeauthor{Audibert:2011} \shortcite{Audibert:2011}, whose basic idea is to truncate the extreme values. Most of the existing work for heavy-tailed bandits develop algorithms based on median of means and truncation \cite{Bubeck:2013,Medina:2016,Han:2018}. 

For heavy-tailed SLB algorithms adopting median of means, it is common to play the chosen arm multiple times and get $r$ payoffs $\{r_{t,a_t}^j\}_{j=1}^r$ at each round. Different ``means'' is considered in existing work \cite{Medina:2016,Han:2018}. The algorithm MoM \cite{Medina:2016} takes the median of $\{r_{t,a_t}^j\}_{j=1}^r$ to conduct least square estimation by one time and the subsequent algorithm MENU \cite{Han:2018} adopts the median of means of least square estimations. More precisely, for $j=1,2,\ldots,r$, the $j$-th estimator in MENU is 
\begin{equation*}
\tilde{\theta}_t^j=\argmin_{\theta\in\R^d}\norm{\widetilde{V}_t\theta-\widetilde{Y}_t^j}^2+\norm{\theta}^2
\end{equation*}
where $\widetilde{V}_t=[x_{\tau,a_\tau}]_{\tau=1}^{t-1}\in\R^{(t-1)\times d}$ and $\widetilde{Y}_t^j=[r_\tau^j]_{\tau=1}^{t-1}\in\R^{(t-1)\times 1}$. After that, the median of means of least square estimations is
\begin{equation*}\label{mom-menu}
m_j= \textnormal{median of }\left\{\norm{\tilde{\theta}_t^j-\tilde{\theta}_t^s}_{\widetilde{A}_t}: s=1,\ldots,r\right\}
\end{equation*}
where $\norm{x}_{\widetilde{A}_t}=\sqrt{x^\top\widetilde{A}_t x}$ for $x\in\R^d$ and $\widetilde{A}_t= I_d+\widetilde{V}_t^\top \widetilde{V}_t$. Then MENU selects the estimator
\begin{equation}\label{esti-menu}
\tilde{\theta}_t^{k_*}{\rm\ where\ }k_*=\argmin_{j\in\{1,2,\ldots,r\}}\{m_j\}
\end{equation}
to predict the payoffs for all arms. 

For heavy-tailed SLB algorithms adopting truncation, the essential difference between existing work is the term chosen to be truncated. The algorithm based on Confidence Region with Truncation (CRT) \cite{Medina:2016} conducts truncation on payoffs $|r_{t,a_t}|$ such that $\tilde{r}_{t,a_t}=r_{t,a_t}\mathbbm{1}_{|r_{t,a_t}|\leq \eta_t}$ for $\eta_t=t^{\frac{1}{2(1+\epsilon)}}$ and obtains the least square estimator through truncated payoffs $\tilde{r}_{t,a_t}$. An improved algorithm TOFU \cite{Han:2018} truncates the term $|u^i_\tau r_{\tau,a_\tau}|$. More precisely, let $[u^1,\ldots,u^d]=\widetilde{A}_t^{-1/2}\widetilde{V}_t^\top$ and $u^i=[u_1^i,u_2^i,\ldots,u_{t-1}^i]$ for $i=1,2,\ldots,d$. The truncation is operated as
\begin{equation*}
\bar{Y}_t^i=[r_{1,a_1}\mathbbm{1}_{|u_1^ir_{1,a_1}|\leq b_t},\ldots,r_{t-1,a_{t-1}}\mathbbm{1}_{|u_{t-1}^ir_{t-1,a_{t-1}}|\leq b_t}]
\end{equation*}
where $b_t=O(t^{\frac{1-\epsilon}{2(1+\epsilon)}})$ and $\mathbbm{1}_{\{\cdot\}}$ is the indicator function. Then the estimator of TOFU is 
\begin{equation}\label{esti-tofu}
\tilde{\theta}_t'=\widetilde{A}_t^{-1/2}[u^1\cdot \bar{Y}_t^1,\ldots,u^d\cdot \bar{Y}_t^d]
\end{equation}
such that $u^i\cdot\bar{Y}_t^i=\sum_{\tau=1}^{t-1}u_\tau^ir_{\tau,a_\tau}\mathbbm{1}_{|u_\tau^ir_{\tau,a_\tau}|\leq b_t}$ for $i=1,2,\ldots,d$.

\section{Algorithms}
In this section, we demonstrate two novel bandit algorithms based on median of means and truncation respectively and illustrate their theoretical guarantees. The detailed proof is provided in supplementary material due to the limitation of space. 

Without loss of generality, we assume feature vectors and target coefficients are contained in the unit ball, that is 
\begin{equation*}
\norm{x_{t,a}}\leq1,\quad\norm{\theta_*} \leq 1.
\end{equation*} 
Following the work of \citeauthor{Chu:2011} \shortcite{Chu:2011}, each of our two original algorithms is divided into basic and master algorithms. The main role of basic algorithms is providing confidence intervals via filtered historical informations, and master algorithm is responsible for ensuring the payoffs' independence.

\subsection{Basic Algorithms}

In the conventional setting where the stochastic payoffs are distributed in $[0,1]$, \citeauthor{Chu:2011} \shortcite{Chu:2011} ultilized the Azuma-Hoeffing's inequality to get the narrow confidence interval \eqref{CI}. Here, we consider the heavy-tailed setting, $\ie$, for some $\epsilon\in(0,1]$, there exists a constant $v>0$, such that
\begin{equation}\label{condition:mom}
\E\left[|r_{t,a_t}-\E[r_{t,a_t}]|^{1+\epsilon}\right]\leq v.
\end{equation}
Note that in this case, Azuma-Hoeffing's inequality is unapplicable as the bounded assumption is violated. The estimator \eqref{LSE} and confidence interval \eqref{CI} are not suitable for heavy-tailed setting. Therefore, the challenge is how to establish a robust estimator associated with proper confidence intervals.

The existing work estimate the payoffs for all arms with a single estimator at each round \cite{Auer:2002,Chu:2011,Medina:2016,Han:2018}, while the expeted payoff $\E[r_{t,a}]$ depends not only on $\theta_*$ but also on the contexts $x_{t,a}$. Thus an intuitive conjecture is that it's better to take estimators adaptive to arms' contexts, and the following example confirms such conjecture.
\begin{exam}\label{example}
We assume $\theta_*=[0.5,0.5]$, the contextual information is $x_{t,1}=[1,0]$ for arm $1$ and $x_{t,2}=[0,1]$ for arm $2$. If we have two estimator $\hat{\theta}_t^1=[0.5,0]$ and $\hat{\theta}_t^2=[0,0.5]$, it's obvious that $\hat{\theta}_t^1$ is a better estimator for $x_{t,1}$ as $x_{t,1}^\top\hat{\theta}_t^1 = x_{t,1}^\top\theta_*$ and $\hat{\theta}_t^2$ is better for $x_{t,2}$.
\end{exam}

The above example encourages us to design estimators adaptive to contexts. 
\subsubsection{Median of Means}
We first present the \underline{b}asic algorithm through \underline{m}edian of \underline{m}eans (BMM) to get confidence intervals for coming payoffs. The complete procedure is provided in Algorithm \ref{BMM}.

To adopt median of means in bandit learning, we play the chosen arm $r$ times and obtain $r$ sequences of payoffs. After that, BMM executes least square estimation for each sequence of payoffs and gets $r$ estimators (Step 1-3). For $j=1,2,\ldots,r$,
\begin{equation}\label{Multi-LSE}
\hat{\theta}_t^j=\argmin_{\theta\in\R^d}\norm{V_t\theta-Y_t^j}^2+\norm{\theta}^2
\end{equation}
where $V_t =[x_{\tau,a_\tau}]_{\tau\in\Psi_t}\in\R^{|\Psi_t|\times d}$ is a matrix of the historical contexts, $Y_t^j=[r_{\tau,a_\tau}^j]_{\tau\in\Psi_t}$ is the historical payoff vector and $\Psi_t\subseteq\{1,2,\ldots,t-1\}$ is an index set filtered by the master algorithm. Then, BMM selects an adaptive estimator $\hat{\theta}_{t,a}$ for each arm $a$ by taking the estimated payoffs as ``means'' (Step 6). More specifically, the estimator for arm $a$ at current round is $\hat{\theta}_{t,a}\in\{\hat{\theta}_t^j\}_{j=1}^r$ such that
\begin{equation}\label{ESTI}
x_{t,a}^\top\hat{\theta}_{t,a}={\rm\ median\ of\ }\{x_{t,a}^\top\hat{\theta}_t^j\}_{j=1}^r.
\end{equation}
By ulilizing median of means, BMM constructs a reliable confidence interval for the expected payoff (Step 6-7), which is 
\begin{equation}\label{BMM-CI}
\left[x_{t,a}^\top\hat{\theta}_{t,a}-w_{t,a},\ x_{t,a}^\top\hat{\theta}_{t,a}+w_{t,a}\right]
\end{equation}
where $w_{t,a}=(\alpha_t+1)\sqrt{x_{t,a}^\top A_tx_{t,a}}, A_t= I_d+V_t^\top V_t$ and $\alpha_t=O(t^{\frac{1-\epsilon}{2(1+\epsilon)}})$. 

\begin{algorithm}[tb]
  \caption{\underline{B}asic algorithm through \underline{M}edian of \underline{M}eans (BMM)}
  \label{BMM}
  \begin{algorithmic}[1]
    \Require
      $\alpha_t\in\R_+,r\in\N,\Psi_t\subseteq\{1,2,\ldots,t-1\}$
    \Ensure
      $\hat{r}_{t,a},\ w_{t,a},\ a=1,2,\ldots,K$
    \State $A_t\gets I_d+\sum_{\tau\in\Psi_t}x_{\tau,a_\tau}x_{\tau,a_\tau}^\top$
    \State $b_t^j\gets \sum_{\tau\in\Psi_t}r_{\tau,a_\tau}^jx_{\tau,a_\tau},\  r_{\tau,a_\tau}^j$ is the $j$-th payoff of playing the arm $a_\tau$ in round $\tau$, $j=1,2,\ldots,r$
    \State $\hat{\theta}_t^j\gets A_t^{-1}b_t^j,\ j=1,2,\ldots,r$
    \State Observe $K$ arm features, $x_{t,1},x_{t,2},\ldots,x_{t,K}\in\R^d$
    \For{ $a=1,2,\ldots,K$}
      \State $\hat{r}_{t,a}\gets x_{t,a}^\top\hat{\theta}_{t,a}$, where $x_{t,a}^\top\hat{\theta}_{t,a}$ is the median of $\{x_{t,a}^\top\hat{\theta}_t^j\}_{j=1}^r$
      \State $w_{t,a}\gets(\alpha_t+1)\sqrt{x_{t,a}^\top A_t^{-1}x_{t,a}}$
    \EndFor
  \end{algorithmic}
\end{algorithm}

When compared with existing algorithms, the main difference lies in how to combine median of means with least square estimation. As we introduced in related work, MoM of \citeauthor{Medina:2016} \shortcite{Medina:2016} and MENU of \citeauthor{Han:2018} \shortcite{Han:2018} take payoffs and the distance between different estimators as ``means'' respectively, while BMM takes estimated payoffs \eqref{ESTI} as ``means'' and predicts coming payoffs with estimators adaptive to contexts. The theoretical guarantee for our estimators is displayed as follows. The payoffs' independence for filtered set $\Psi_t$ is ensured by the master algorithm SupBMM and we will present it later.

\begin{prop}\label{prop:1}
For fixed feature vectors $x_{\tau,a_\tau}$ with $\tau\in\Psi_t$ in BMM, the payoffs $\{r^j_{\tau,a_\tau}\}_{\tau\in\Psi_t}, j=1,2,\ldots,r$ are independent random variables which satisfy \eqref{expected_payoff} and \eqref{condition:mom}. Then, if $\alpha_t=(12v)^{\frac{1}{1+\epsilon}}t^{\frac{1-\epsilon}{2(1+\epsilon)}}$ and  $r=\left \lceil 8\ln\frac{2KT\ln T}{\delta} \right \rceil$, with probability at least $1-\delta/T$, for any $a\in\{1,2,\ldots,K\}$, we have
\begin{equation*}
|\hat{r}_{t,a}-x_{t,a}^\top\theta_*|\leq(\alpha_t+1)\sqrt{x_{t,a}^\top A_t^{-1}x_{t,a}}.
\end{equation*}
\end{prop}

\begin{rem}\label{rem:1}
\textnormal{
The confidence interval of BMM depends on the $1+\epsilon$ central moment of the payoff distribution, which is constructed at the cost of $r$ times to play the chosen arm. When the payoffs admit a finite variance, i.e., $\epsilon=1$, our algorithm utilizes tighter confidence intervals with $\alpha_t=\sqrt{12v}$, in contrast, \citeauthor{Chu:2011} \shortcite{Chu:2011} constructed confidence intervals with $\alpha_t=O(\sqrt{\ln(TK)})$.
}
\end{rem}

\subsubsection{Truncation}
In this section, we develop the \underline{b}asic algorithm through \underline{t}run\underline{c}ation (BTC) to get confidence intervals for coming payoffs. The complete procedure is provided in Algorithm \ref{BTC}.

For heavy-tailed SLB algorithms adopting truncation, the key point is how to combine the least square estimation with truncation. The existing least square estimator \eqref{LSE} without truncation does not take use of current epoch's contexts $x_{t,a}$, while Example \ref{example} encourages us to consider adaptive estimator. The estimated payoff of \citeauthor{Chu:2011} \shortcite{Chu:2011} is a linear combination of historical payoffs, $\ie$,
\begin{equation*} 
x_{t,a}^\top A_t^{-1}V_t^\top Y_t=\sum_{\tau\in\Psi_t}\beta_\tau r_{\tau,a_\tau}
\end{equation*}
where $x_{t,a}^\top A_t^{-1}V_t^\top=[\beta_\tau]_{\tau\in\Psi_t}\in\R^{1\times|\Psi_t|}$ depending on contexts. For the sake of designing an estimator adaptive to contexts, BTC truncates term $\beta_\tau r_{\tau,a_\tau}$ (Step 7) and obtains estimated payoff,
\begin{equation}
x_{t,a}^\top A_t^{-1}V_t^\top \widehat{Y}_t=\sum_{\tau\in\Psi_t}\beta_\tau r_{\tau,a_\tau}\mathbbm{1}_{|\beta_\tau r_{\tau,a_\tau}|\leq h_{t,a}}
\end{equation}
where $\widehat{Y}_t=\left[r_{\tau,a_\tau}\mathbbm{1}_{|\beta_\tau r_{\tau,a_\tau}|\leq h_{t,a}}\right]_{\tau\in\Psi_t}$ and $h_{t,a}$ is the truncation criterion. We set $h_{t,a}=\norm{x_{t,a}^\top A_t^{-1}V_t^\top}_{1+\epsilon}$, and the confidence interval for the adaptive estimator $\hat{\theta}_{t,a}=A_t^{-1}V_t^\top\widehat{Y}_t$ is 
\begin{equation}\label{CI-trunc}
\left[x_{t,a}^\top\hat{\theta}_{t,a}-w_{t,a},\ x_{t,a}^\top\hat{\theta}_{t,a}+w_{t,a}\right]
\end{equation}
where $w_{t,a}=(\alpha_t+1)\sqrt{x_{t,a}^\top A_t^{-1}x_{t,a}}$ and $\alpha_t=O(\ln(TK)t^{\frac{1-\epsilon}{2(1+\epsilon)}})$ (Step 9-10).

\begin{algorithm}[tb]
  \caption{\underline{B}asic algorithm through \underline{T}run\underline{c}ation (BTC)}
  \label{BTC}
  \begin{algorithmic}[1]
    \Require
      $\alpha_t\in\R_+, \ \Psi_t\subseteq\{1,2,\ldots,t-1\}$
    \Ensure
      $\hat{r}_{t,a},\ w_{t,a}, \  a=1,2,\ldots,K$
    \State $A_t\gets I_d+\sum_{\tau\in\Psi_t}x_{\tau,a_\tau}x_{\tau,a_\tau}^\top$
    \State $V_t \gets [x_{\tau,a_\tau}]_{\tau\in\Psi_t}$
    \State Observe $K$ arm features,\ $x_{t,1},x_{t,2},\ldots,x_{t,K}\in\R^d$
    \For{ $a=1,2,\ldots,K$}
      \State $[\beta_{\tau_1},\beta_{\tau_2},\ldots,\beta_{\tau_{|\Psi_t|}}]\gets x_{t,a}^\top A_t^{-1}V_t^\top$
      \State $h_{t,a}\gets\norm{x_{t,a}^\top A_t^{-1}V_t^\top}_{1+\epsilon}$
      \State $\widehat{Y}_{t,a}\gets[\hat{r}_{\tau,a_\tau}]_{\tau\in\Psi_t}$  where $\hat{r}_{\tau,a_\tau}=r_{\tau,a_\tau}\mathbbm{1}_{|\beta_\tau r_{\tau,a_\tau}|\leq h_{t,a}}$
      \State $\hat{\theta}_{t,a}\gets A_t^{-1}V_t^\top\widehat{Y}_{t,a}$
      \State $\hat{r}_{t,a}\gets x_{t,a}^\top\hat{\theta}_{t,a}$
      \State $w_{t,a}\gets(\alpha_t+1)\sqrt{x_{t,a}^\top A_t^{-1}x_{t,a}}$
    \EndFor
  \end{algorithmic}
\end{algorithm}

When compared with existing work, the main difference lies in the term chosen to be truncated. CRT of \citeauthor{Medina:2016} \shortcite{Medina:2016} truncates the payoff $r_{t,a_t}$ and TOFU of \citeauthor{Han:2018} \shortcite{Han:2018} truncates the term $u_\tau^ir_{\tau,a_\tau}$ as we mentioned in related work. Since $r_{t,a_t}$ and $u_\tau^ir_{\tau,a_\tau}$ do not depend on current epoch's contexts, the estimators of CRT and TOFU are not adaptive. BTC develops an adaptive estimator $\hat{\theta}_{t,a}$ by performing least square estimation with the truncated term $\beta_\tau r_{\tau,a_\tau}$. Whether the confidence interval \eqref{CI-trunc} is true is the main difficulty in the analysis of estimator $\hat{\theta}_{t,a}$ because truncation results in a bias.

BTC requires that for some $\epsilon\in(0,1]$, the $1+\epsilon$ raw moment of the payoffs is bounded, $\ie$, there is a constant $v>0$, the payoffs admit
\begin{equation}\label{condition:trunc}
\E\left[|r_{t,a_t}|^{1+\epsilon}\right]\leq v.
\end{equation}

\begin{prop}\label{prop:2}
For fixed feature vectors $x_{\tau,a_\tau}$ with $\tau\in\Psi_t$ in BTC, the payoffs $\{r_{\tau,a_\tau}\}_{\tau\in\Psi_t}$ are independent random variables which satisfy \eqref{expected_payoff} and \eqref{condition:trunc}. If $\alpha_t=\left(\frac{2}{3}\ln\frac{2TK\ln T}{\delta}+\sqrt{2\ln\frac{2TK\ln T}{\delta}v}+v\right)t^{\frac{1-\epsilon}{2(1+\epsilon)}}$, then with probability at least $1-\delta/T,\  \forall a\in\{1,2,\ldots,K\}$, we have 
\begin{equation*}
|\hat{r}_{t,a}-x_{t,a}^\top\theta_*|\leq(\alpha_t+1)\sqrt{x_{t,a}^\top A_t^{-1}x_{t,a}}.
\end{equation*}
\end{prop}

\begin{rem}\textnormal{
The above proposition indicates that the confidence interval \eqref{CI-trunc} provided by BTC is true wih high probability. BTC is less expensive when compared with TOFU of \citeauthor{Han:2018} \shortcite{Han:2018} as $A_t^{-1}V_t^\top$ can be computed online by the Sherman-Morrison formula \cite{Matrix-computations} while $\widetilde{A}_t^{-1/2}$ of TOFU can not. As we mentioned in related work, TOFU has to store both historical contextual matrix $\widetilde{V}_t$ and historical payoffs $\{r_{\tau,a_\tau}\}_{\tau=1}^{t-1}$, while BTC only needs to store historical payoffs.
}
\end{rem} 

\subsection{Master Algorithm}\label{sec:Mas-Alg}
In this section, we demonstrate the master algorithm to settle the independence issue and establish its theoretical gurantees. The master algorithm is adapted from SupLinUCB \cite{Chu:2011} and the complete procedure is summarized in Algorithm \ref{Mas-Alg}.

The master algorithm is responsible for ensuring the payoffs' independence in $\Psi_t$ as well as achieving the trade-off between exploitation and exploration through the confidence intervals provided by basic algorithms. At round $t$, the algorithm screens the candidate arms through $S$ stages until an arm is chosen. The algorithm chooses an arm either when the expected payoff is close to the optimal one or when the confidence interval's width is large. More precisely, we consider three situations at each stage.

If the estimation payoffs of all arms are accurate enough, which means the confidence level is up to $1/\sqrt{T}$ (Step 8), we do not need to do exploration and choose the arm maxmizing the upper confidence bound. Otherwise, we notice that the width of confidence interval at stage $s$ is supposed to be $2^{-s}$. If $w_{t,a_t}^s>2^{-s}$ for some $a_t\in\hat{A}_s$ (Step 11), we play it to take more exploration on this arm. The last situation is that we can not decide which arm to choose at current stage (Step 13), and only those arms which are sufficiently close to the optimal arm are filtered to the next stage (Step 14). When the width of all confidence intervals is at most $2^{-s}$ and the arm $a$ satisfies $\hat{r}_{t,a}^s+w_{t,a}^s<\hat{r}_{t,a'}^s+w_{t,a'}^s-2^{1-s}$ for some $a'\in\hat{A}_{s}$, then the arm $a$ cannot be optimal and does not pass to the next stage. The master algorithms taking BMM and BTC as subroutines are called SupBMM and SupBTC, respectively.

\begin{algorithm}[tb]
  \caption{Master Algorithm (SupBMM and SupBTC)}
  \label{Mas-Alg}
  \begin{algorithmic}[1]
    \Require
        $T\in\mathbb{N}$      
    \State $S\gets\lfloor\ln T\rfloor$
    \State $\Psi_1^s\gets\varnothing$ for all $s\in\{1,2,\ldots,S\}$
    \For{ $t=1,2,\ldots,T$}
        \State $s\gets1, \hat{A}_1\gets\{1,2,\ldots,K\}$
        \Repeat
          	\State \textbf{SupBMM:} Use BMM with $\Psi_t^s$ to calculate the width $w_{t,a}^s$ and upper confidence bound $\hat{r}_{t,a}^s+w_{t,a}^s$ for every $a\in\hat{A}_s$
    		\State \textbf{SupBTC:} Use BTC with $\Psi_t^s$ to calculate the width $w_{t,a}^s$ and upper confidence bound $\hat{r}_{t,a}^s+w_{t,a}^s$ for every $a\in\hat{A}_s$
            \If{ $w_{t,a}^s\leq1/\sqrt{T} \ \ \  \forall a\in\hat{A}_s $ }
                \State Choose $a_t=\argmax_{a\in{\hat{A}_s}}(\hat{r}_{t,a}^s+w_{t,a}^s)$
                \State Keep the same index sets at all levels: $\Psi_{t+1}^{s'}\gets\Psi_{t}^{s'}\ \  \forall s'\in\{1,2,\ldots,S\}$
            \ElsIf  { $w_{t,a_t}^s>2^{-s}$ for some $a_t\in\hat{A}_s$ } 
                \State Choose this arm $a_t$ and update the index sets at all levels:
\begin{equation*}
\Psi_{t+1}^{s'}\gets
\begin{cases}
\Psi_t^{s'}\cup\{t\}  &\text{if} \ s'=s \\
\Psi_t^{s'} &\text{otherwise}
\end{cases}
\end{equation*}
            \Else { $w_{t,a}^s\leq 2^{-s}\ \ \forall a\in\hat{A}_s$ }
                \State $\hat{A}_{s+1} \gets \{ a\in\hat{A}_s | \hat{r}_{t,a}^s+w_{t,a}^s\geq \max_{a'\in\hat{A}_s}(\hat{r}_{t,a'}^s+w_{t,a'}^s)-2^{1-s}\}$
                \State $s\gets s+1$
            \EndIf
        \Until {an arm $a_t$ is found.}
        \State \textbf{SupBMM:} Play $a_t$ $r$ times and observe payoffs $r_{t,a_t}^1,r_{t,a_t}^2,\ldots,r_{t,a_t}^r$
        \State \textbf{SupBTC:} Play $a_t$ and observe payoff $r_{t,a_t}$
    \EndFor
  \end{algorithmic}
\end{algorithm}

For the independence, we notice that the updation of $\Psi_t^s$ is associated with the historical trails in set $\bigcup_{\sigma<s}\Psi^\sigma_t$ and $w^s_{t,a}$, while $w^s_{t,a}$ only depends on contexts $x_{t,a}$ and $x_{\tau,a_\tau}$ with $\tau\in\Psi_{t}^{s}$. Thus the payoffs $r_{\tau,a_\tau}$ are independent random variables for any fixed sequence of $x_{\tau,a_\tau}$ with $\tau\in\Psi_{t}^{s}$. Proposition \ref{prop:1} and Proposition \ref{prop:2} are established with payoffs' independence, which leads to the regret bounds of SupBMM and SupBTC.

\begin{thm}\label{thm:1}
Assume all payoffs admit \eqref{expected_payoff} and \eqref{condition:mom}. Let $r=\left \lceil 8\ln\frac{2KT\ln T}{\delta} \right \rceil$, for any $\delta\in(0,1)$, with probability at least $1-\delta$, the regret of SupBMM satisfies 
\begin{equation*}
\begin{aligned}
R(T)\leq &120\left(1+(12v)^{\frac{1}{1+\epsilon}}\right)\ln T\sqrt{2d\ln\frac{2KT\ln T}{\delta}}T^{\frac{1}{1+\epsilon}}\\
 &+5\sqrt{2T\ln\frac{2KT\ln T}{\delta}}.
\end{aligned}
\end{equation*}
\end{thm}

\begin{rem}\textnormal{We achieve a regret bound of order $\widetilde{O}(d^{\frac{1}{2}}T^{\frac{1}{1+\epsilon}})$. For $\epsilon=1$, it reduces to an $\widetilde{O}(\sqrt{dT})$ bound, which implies that we get the same order as bounded payoffs assumption in terms of both $d$ and $T$ \cite{Chu:2011}. We point out that for any random variable $X$,
\begin{equation*}
\E[|X-\E X|^{\epsilon_1}]\leq\left(\E[|X-\E X|^{\epsilon_2}]\right)^{\frac{\epsilon_1}{\epsilon_2}}
\end{equation*}
where $\epsilon_1,\epsilon_2\in\R_+$ and $\epsilon_1\leq\epsilon_2$. Therefore, our upper bound $\widetilde{O}(\sqrt{dT})$ also holds for the payoffs with finite higher order ($\epsilon>1$) central moments, and matches the lower bound of bounded payoffs up to some polylogarithmic factors. 
}
\end{rem}

\begin{thm}\label{thm:2}
Assume all payoffs admit \eqref{expected_payoff} and \eqref{condition:trunc}. For any $\delta\in(0,1)$, with probability at least $1-\delta$, the regret of SupBTC satisfies
\begin{equation*} 
\begin{aligned}
R(T)\leq&2\sqrt{T}+40\ln T\sqrt{dT}+40\left(\frac{2}{3}\ln\frac{2TK\ln T}{\delta}+\right.\\
 &\left.\sqrt{2\ln\frac{2TK\ln T}{\delta}v}+v\right)\ln T\sqrt{d} T^{\frac{1}{1+\epsilon}}.
\end{aligned}
\end{equation*}
\end{thm}

\begin{rem}\textnormal{The above theorem assumes a finite $1+\epsilon$ raw moment of payoffs and achieves a regret bound of the same order $\widetilde{O}(d^{\frac{1}{2}}T^{\frac{1}{1+\epsilon}})$ as Theorem \ref{thm:1}, while Theorem \ref{thm:1} depends on the $1+\epsilon$ central moments of payoffs. \citeauthor{Han:2018} \shortcite{Han:2018} proposed the algorithms achieving the regret bound $\widetilde{O}(dT^{\frac{1}{1+\epsilon}})$, so our algorithms have a better dependence on $d$ for finite-armed SLB with heavy-tailed payoffs.
}
\end{rem}

\section{Lower Bound}
In this section, we give the lower bound for finite-armed SLB with heavy-tailed payoffs.

\begin{thm}\label{lower bound:SLB}
For any algorithm $\A$ with $T\geq K\geq4$ and $T\geq(2d)^{\frac{1+\epsilon}{2\epsilon}}$, let $\gamma=(K/(T+2K))^{\frac{1}{1+\epsilon}}$, there exists a sequence of feature vectors $\{x_{t,a}\}_{t=1}^T$ for $a=1,2,\ldots,T$ and a coefficient vector $\theta_*$ such that the payoff for each arm is in $\{0,1/\gamma\}$ with mean $x_{t,a}^\top\theta_*$. If $d\geq K$, we have
\begin{equation*}
\E\left[R(T)\right]\geq\frac{1}{32}\left(d-1\right)^{\frac{\epsilon}{1+\epsilon}}T^{\frac{1}{1+\epsilon}}= O\left(d^{\frac{\epsilon}{1+\epsilon}}T^{\frac{1}{1+\epsilon}}\right).
\end{equation*}
\end{thm}

\begin{rem}\textnormal{
The above theorem essentially establishes an $\Omega(d^{\frac{\epsilon}{1+\epsilon}}T^{\frac{1}{1+\epsilon}})$ lower bound associated with $d$ and $T$ for SLB under the heavy-tailed setting, which matches  the upper bounds of Theorems \ref{thm:1} and \ref{thm:2} in the sense of the polynomial order on $T$. To the best of our knowledge, this is the first lower bound for finite-armed SLB with heavy-tailed payoffs. 
}
\end{rem}
\section{Analysis}
Although our analysis is built based on the results of \citeauthor{Auer:2002} \shortcite{Auer:2002}, we have further slacken the payoffs to be heavy-tailed distributions. For the upper bound of SupBTC, the key point is to deduce a narrow confidence interval with high probability as we do in Propositions 2. For the lower bound, we demonstrate a lower bound for the muti-armed bandits (MAB) with heavy-tailed payoffs first, and then extend it to stochastic linear bandits (SLB) by a proper design of the contextual feature vectors $x_{t,a}$ and coefficient vector $\theta_*$. To begin with the analysis, we define a new term 
\[
c_{t,a} =\sqrt{x_{t,a}^\top A_t^{-1}x_{t,a}}
\]
for the simplification, and the following assumptions on payoffs is needed. 

\begin{ass}\label{ass:1}
For some $\epsilon\in(0,1]$, there exists a constant $v>0$, such that
\begin{equation*}\label{condition:mom}
\E\left[|r_{t,a_t}^j-\E[r_{t,a_t}^j]|^{1+\epsilon}\right]\leq v
\end{equation*}
\end{ass}

\begin{ass}\label{ass:2}
For fixed feature vectors $x_{\tau,a_\tau}$ with $\tau\in\Psi_t$ in BMM, the payoffs $\{r^j_{\tau,a_\tau}\}_{\tau\in\Psi_t}, j=1,2,\ldots,r$ are independent random variables with $\E[r^j_{\tau,a_\tau}]=x_{\tau,a_\tau}^T\theta_*$.
\end{ass}

\subsection{Proof of Proposition 1}
The latent idea in BMM is median of means and it's necessary to give a theoretical guarantee for each ``mean" first. We will display such guarantee in Lemma \ref{lem:3} and before this, we introduce the following tool which can be used to conduct Lemma \ref{lem:3}.

\begin{lem}\label{lem:2}
Let $X_1,\ldots ,X_n$ be independent random variables with mean $\E[X_i]=0$, and $\E[|X_i|^{1+\epsilon}]\leq v$. Then for some fixed numbers $\beta_1,\beta_2,\ldots,\beta_n\in\R$ and $\chi>0$ ,
\begin{equation*}
\begin{aligned}
 &\mathbb{P}\left\{\left|\sum_{i=1}^{n}\beta_iX_i\right|>\chi\right\}\\
\leq&2\sum_{i=1}^{n}|\beta_i|^{1+\epsilon}\frac{v}{\chi^{1+\epsilon}}+\left(\sum_{i=1}^{n}|\beta_i|^{1+\epsilon}\frac{v}{\chi^{1+\epsilon}}\right)^2.
\end{aligned}
\end{equation*}
\end{lem}

 We point out that Lemma \ref{lem:2} is a variant of Chebyshev inequality and detailed proof is provided at the end of the analysis section.

\begin{lem}\label{lem:3}
For fixed feature vectors $x_{\tau,a_\tau}$ with $\tau\in\Psi_t$ in BMM, the payoffs $\{r^j_{\tau,a_\tau}\}_{\tau\in\Psi_t}$ are independent random variables satisfying Assumptions \ref{ass:1} and \ref{ass:2}. For $a=1,2,\ldots,K$ and $j=1,2,\ldots,r$, we have
\begin{equation*}
\mathbb{P}\left\{\left|x_{t,a}^\top A_t^{-1}V_t^\top(Y_t^j-V_t\theta_*)\right|>(12v)^{\frac{1}{1+\epsilon}}t^{\frac{1-\epsilon}{2(1+\epsilon)}} c_{t,a}\right\}\leq\frac{1}{4}.
\end{equation*}
\end{lem}

\noindent\textbf{Proof} First of all, we display a fact, 
\begin{equation}\label{fact}
\begin{split}
c_{t,a}=&\left(x_{t,a}^\top A_t^{-1}x_{t,a}\right)^\frac{1}{2}\\
=&\left(x_{t,a}^\top A_t^{-1}(I_d+V_t^\top V_t)A_t^{-1}x_{t,a}\right)^\frac{1}{2}\\
\geq&\left(x_{t,a}^\top A_t^{-1}V_t^\top V_tA_t^{-1}x_{t,a}\right)^\frac{1}{2}\\
=&\norm{x_{t,a}^\top A_t^{-1}V_t^\top }\\
\geq&t^{\frac{\epsilon-1}{2(1+\epsilon)}}\norm{ x_{t,a}^\top A_t^{-1}V_t^\top }_{1+\epsilon}
\end{split}
\end{equation}
where the last inequality is due to H$\ddot{o}$lder inequality. For $j=1,2,\ldots,r$, it's easy to verify that
$$\E\left[Y_t^j-V_t \theta_*\right]=0.$$
According to Lemma \ref{lem:2}, we have 
\begin{align*}
&\mathbb{P}\left\{\left|x_{t,a}^\top A_t^{-1}V_t^\top (Y_t^j-V_t\theta_*)\right|>(12v)^{\frac{1}{1+\epsilon}}t^{\frac{1-\epsilon}{2(1+\epsilon)}}  c_{t,a}\right\}\\
\leq&\frac{\norm{ x_{t,a}^\top A_t^{-1}V_t^\top }_{1+\epsilon}^{1+\epsilon}}{6t^\frac{1-\epsilon}{2}c_{t,a}^{1+\epsilon}}+\left(\frac{\norm{ x_{t,a}^\top A_t^{-1}V_t^\top }_{1+\epsilon}^{1+\epsilon}}{12t^\frac{1-\epsilon}{2}c_{t,a}^{1+\epsilon}}\right)^2\\
\leq&\frac{\norm{ x_{t,a}^\top A_t^{-1}V_t^\top }_{1+\epsilon}^{1+\epsilon}}{4t^\frac{1-\epsilon}{2}c_{t,a}^{1+\epsilon}}.
\end{align*}
The fact \eqref{fact} tells us
\begin{align*}
\mathbb{P}\left\{\left|x_{t,a}^\top A_t^{-1}V_t^\top (Y_t^j-V_t\theta_*)\right|>(12v)^{\frac{1}{1+\epsilon}}t^{\frac{1-\epsilon}{2(1+\epsilon)}}c_{t,a}\right\}&\leq\frac{1}{4}.
\end{align*}
$\hfill\blacksquare$

An intuitive explanation for Lemma \ref{lem:3} is more than a half of the estimated payoffs for arm $a$ is bounded by the upper confidence bound in expectation.

\noindent\textbf{Proof of Proposition 1} 
For $a=1,2,\ldots,K$, we have
\begin{equation}\label{equation:prop1}
\begin{aligned}
 &\hat{r}_{t,a}-x_{t,a}^*\\
=&x_{t,a}^\top \hat{\theta}_{t,a}-x_{t,a}^\top \theta_*\\
=&x_{t,a}^\top A_t^{-1}b_{t,a}-x_{t,a}^\top A_t^{-1}(I_d+V_t^\top V_t)\theta_*\\
=&x_{t,a}^\top A_t^{-1}V_t^\top Y_{t,a}-x_{t,a}^\top A_t^{-1}(\theta_*+V_t^\top V_t\theta_*)\\
=&x_{t,a}^\top A_t^{-1}V_t^\top (Y_{t,a}-V_t\theta_*)-x_{t,a}^\top A_t^{-1}\theta_*
\end{aligned} 
\end{equation}
where $Y_{t,a}$ is the payoff vector used to calculate $\hat{\theta}_{t,a}$ such that $x_{t,a}^\top A_t^{-1}V_t^\top Y_{t,a}$ is the median of $\{x_{t,a}^\top A_t^{-1}V_t^\top Y_t^j\}_{j=1}^r$. With $\norm{\theta_*}\leq 1$, we have
\begin{equation}\label{equation:1}
\begin{aligned}
|\hat{r}_{t,a}-x_{t,a}^*|\leq&|x_{t,a}^\top A_t^{-1}V_t^\top (Y_{t,a}-V_t\theta_*)|\\
 &+\norm{A_t^{-1}x_{t,a}}.
\end{aligned}
\end{equation} 
The second term of the right-hand side in inequality \eqref{equation:1} is bounded by the following fact:
\begin{equation}\label{lem2:right2}
\begin{split}
\norm{A_t^{-1}x_{t,a}}_2=&\sqrt{x_{t,a}^\top A_t^{-1}I_dA_t^{-1}x_{t,a}}\\
\leq&\sqrt{x_{t,a}^\top A_t^{-1}(I_d+V_t^\top V_t)A_t^{-1}x_{t,a}}\\
=&\sqrt{x_{t,a}^\top A_t^{-1}x_{t,a}}\\
 =&c_{t,a}.
\end{split}
\end{equation}
In order to give a bound on the first term of right-hand side in inequality \eqref{equation:1}, we consider the Lemma \ref{lem:3} and the independence of samples indexed in $\Psi_t$. For $j=1,2,\ldots,r$, we have
\begin{equation*}
\mathbb{P}\left\{\left|x_{t,a}^\top A_t^{-1}V_t^\top (Y_t^j-V_t\theta_*)\right|>\alpha_t c_{t,a}\right\}\leq\frac{1}{4}
\end{equation*}
where $\alpha_t=(12v)^{\frac{1}{1+\epsilon}}t^{\frac{1-\epsilon}{2(1+\epsilon)}}$.  We define the random variables
\begin{equation*}
\begin{aligned}
X_j=\mathbbm{1}_{x_{t,a}^\top A_t^{-1}V_t^\top(Y_t^j-V_t\theta_*)>\alpha_t c_{t,a} }
\end{aligned}
\end{equation*}
so that $p_j=\mathbb{P}\{X_j=1\}\leq\frac{1}{4}$. From the Azuma-Hoeffing's inequality \cite{azuma1967}, 
\begin{align*}
\mathbb{P}\left\{\sum_{j=1}^{r}X_j\geq\frac{r}{2}\right\}\leq&\mathbb{P}\left\{\sum_{j=1}^{r}X_j-p_j\geq \frac{r}{4}\right\}\\
\leq &e^{-r/8}\leq\frac{\delta}{2KT}
\end{align*}
for $r=\left \lceil 8\ln\frac{2KT\ln T}{\delta} \right \rceil$. The inequality $\sum_{j=1}^{r}X_j\geq \frac{r}{2}$ means more than half of the terms $\{X_j\}_{j=1}^{r}$ is true. Thus, the median term $x_{t,a}^\top A_t^{-1}V_t^\top Y_{t,a}$ satisfies 
$$x_{t,a}^\top A_t^{-1}V_t^\top(Y_{t,a}-V_t\theta^*)>\alpha_t c_{t,a}$$ 
with probability at most $\frac{\delta}{2KT}$. A similar argument shows that $x_{t,a}^\top A_t^{-1}V_t^\top(Y_{t,a}-V_t\theta^*)<-\alpha_t c_{t,a}$ with probability at most $\frac{\delta}{2KT}$. Therefore, we have
\begin{equation*}
|x_{t,a}^\top A_t^{-1}V_t^\top(Y_{t,a}-V_t\theta^*)|<\alpha_t c_{t,a}
\end{equation*}
with probability at least $1-\frac{\delta}{KT}$.

Now, using a union bound with respect to all arms deduces that with probability at least $1-\delta/T$, for any $a\in\{1,2,\ldots,K\}$, we have
\begin{equation}\label{lem2:right1}
|x_{t,a}^\top A_t^{-1}V_t^\top (Y_{t,a}-V_t\theta_*)|<\alpha_t c_{t,a}.
\end{equation}
Combining the inequality \eqref{equation:1}, inequality \eqref{lem2:right2} and inequality \eqref{lem2:right1} completes the proof.
$\hfill\blacksquare$

\subsection{Proof of Theorem 1}

\begin{lem}\label{lem:4}\textnormal{(\cite{Chu:2011}, Lemma 3)}
With the notations of BMM and assuming $|\Psi^s_{T+1}|\geq2$ for $s=1,2,\ldots,S$, we have 
\begin{equation*}
\sum_{t\in\Psi^s_{T+1}}c_{t,a_t}\leq 5\sqrt{d|\Psi^s_{T+1}|\ln|\Psi^s_{T+1}|}.
\end{equation*}
\end{lem}

\begin{lem}\label{lem:1}\textnormal{(\cite{Auer:2002}, Lemma 14)}
For each $s=1,2,\ldots,S,\ \ t=1,2,\ldots,T$ and any fixed sequence of arms $x_{\tau,a_{\tau}}$ with $\tau\in\Psi_{t}^{s}$, the payoffs $\{r_{\tau,a_\tau}\}_{\tau\in\Psi_t}$ are independent random variables with $\E[r_{\tau,a_\tau}]=x_{\tau,a_\tau}^\top\theta_*$. 
\end{lem}

\begin{lem}\label{lem:5}\textnormal{(\cite{Auer:2002}, Lemma 15)}
With probability $1-\delta S$, for any $t=1,2,\ldots,T$ and $s=1,2,\ldots,S$,
\begin{align*}
|\hat{r}_{t,a}-\E[r_{t,a}]|\leq& w_{t,a}\ \ \forall a\in\{1,2,\ldots,K\},\\
a_t^*\in&\hat{A}_s \ and\\
\E[r_{t,a_t^*}]-\E[r_{t,a}]\leq&2^{3-s}\ \ \forall a\in \hat{A}_s.
\end{align*}
\end{lem}

\begin{lem}\label{lem:6}
If we set $\alpha=(12v)^{\frac{1}{1+\epsilon}}T^{\frac{1-\epsilon}{2(1+\epsilon)}}$, for all $s=1,2,\ldots,S$,
\begin{equation*}
|\Psi^s_{T+1}|\leq 5\times2^s(1+\alpha)\sqrt{d|\Psi^s_{T+1}|\ln|\Psi^s_{T+1}|}.
\end{equation*}
\end{lem}

\noindent\textbf{Proof}
From Lemma \ref{lem:4}, we have
\begin{equation*}
\begin{split}
\sum_{\tau\in\Psi^s_{T+1}}w_{\tau,a_\tau}^s=&\sum_{\tau\in\Psi^s_{T+1}}(1+\alpha_t)c_{\tau,a_\tau}\\
\leq&5(1+\alpha)\sqrt{d|\Psi^s_{T+1}|\ln|\Psi^s_{T+1}|}.
\end{split}
\end{equation*}
By the second case of SupBMM (else if case), we get
$$\sum_{\tau\in\Psi^s_{T+1}}w_{\tau,a_\tau}^s\geq2^{-s}|\Psi^s_{T+1}|.$$
Obviously, 
$$|\Psi^s_{T+1}|\leq 5\times2^s(1+\alpha)\sqrt{d|\Psi^s_{T+1}|\ln|\Psi^s_{T+1}|}.$$
We have finished the proof.
$\hfill\blacksquare$

Lemma \ref{lem:5} provides a more precise description for the arms in different stages and Lemma \ref{lem:6} tells us $|\Psi^s_{T+1}|$ can be bounded by $\sqrt{|\Psi^s_{T+1}|\ln|\Psi^s_{T+1}|}$, which means $|\Psi^s_{T+1}|$ can be successfully bounded by a term with lower order.

\noindent\textbf{Proof of Theorem 1}
When the basic algorithm is BMM, the chosen arm plays $r$ times at each epoch in algorithm SupBMM. Therefore, we set $T_0=\lfloor T/r \rfloor$ and the master algorithm plays $T_0$ rounds. From the procedure of SupBMM, we have 
\[
\bigcup_{s\in\{1,2,\ldots,S\}}\Psi_{T_0+1}^s\subset\{1,2,\ldots,T_0\}.
\]
If we set $\Psi_0=\{1,2,\ldots,T_0\}\setminus\bigcup_{s\in\{1,2,\ldots,S\}}\Psi_{T_0+1}^s$, for any $t\in\Psi_0$,
\begin{equation*}
\begin{aligned}
\E[r_{t,a_t^*}]-\E[r_{t,a_t}]\leq&\E[r_{t,a_t^*}]-(x_{t,a_t^*}^\top\hat{\theta}_{t,a_t^*}+w_{t,a_t^*})\\
 &+(x_{t,a_t}^\top\hat{\theta}_{t,a_t}+w_{t,a_t})-\E[r_{t,a_t}]\\
\leq&2w_{t,a_t}\leq\frac{2}{\sqrt{T_0}}.
\end{aligned}
\end{equation*}
From the Lemma \ref{lem:5} and Lemma \ref{lem:6}, with probability at least $1-S\delta$, where $S=\lfloor \ln T_0 \rfloor$, we have
\begin{equation*}
\begin{split}
R(T_0)=&\sum_{t=1}^{T_0}\E[r_{t,a_t^*}]-\E[r_{t,a_t}]\\
=&\sum_{t\in\Psi_0}\left(\E[r_{t,a_t^*}]-\E[r_{t,a_t}]\right)\\
 &+\sum_{s=1}^S\sum_{t\in\Psi^s_{T_0+1}}\left(\E[r_{t,a_t^*}]-\E[r_{t,a_t}]\right)\\
\leq&\frac{2}{\sqrt{T_0}}|\Psi_0|+\sum_{s=1}^S8\times2^{-s}|\Psi^s_{T_0+1}|\\
\leq&\frac{2}{\sqrt{T_0}}|\Psi_0|+40(1+\alpha)\sum_{s=1}^S\sqrt{d|\Psi^s_{T_0+1}|\ln|\Psi^s_{T+1}|}\\
\leq&2\sqrt{T_0}+40(1+\alpha)\sqrt{Sd}\sqrt{\sum_{s=1}^S|\Psi^s_{T_0+1}|\ln|\Psi^s_{T_0+1}|}\\
\leq&2\sqrt{T_0}+40(1+\alpha)\ln T_0\sqrt{dT_0}.
\end{split}
\end{equation*}
Through the relationship $R(T)=rR(T_0)$, we have
$$R(T)\leq2\sqrt{Tr}+40(1+\alpha)\ln T_0\sqrt{dTr}$$
with probability at least $1-\delta S$. 

Replace $\delta$ with $\delta/\ln T$ and take $r=\lceil8\ln\frac{2KT\ln T}{\delta}\rceil, \alpha=(12v)^{\frac{1}{1+\epsilon}}T^{\frac{1-\epsilon}{2(1+\epsilon)}}$ into the above inequality, we have
\begin{equation*}
\begin{split}
R(T)\leq& 5\sqrt{2T\ln\frac{2KT\ln T}{\delta}}\\
 &+120(1+\alpha)\ln T_0\sqrt{2dT\ln\frac{2KT\ln T}{\delta}}\\
=&5\sqrt{2T\ln\frac{2KT\ln T}{\delta}}+\\
 &120\left(1+(12v)^{\frac{1}{1+\epsilon}}\right)\ln T\sqrt{2d\ln\frac{2KT\ln T}{\delta}}T^{\frac{1}{1+\epsilon}}
\end{split}
\end{equation*}
with probability at least $1-\delta$.

\subsection{Proof of Proposition 2}

There are some similiar techniques between Propositions 1 and 2 since the goal of BMM and BTC is estimating payoffs for current epoch. To simplify the proof, the common parts such as inequalities \eqref{equation:prop1}, \eqref{equation:1}, \eqref{lem2:right2} are omitted and what we need to bound is the term $|x_{t,a}^\top A_t^{-1}V_t^\top (\widehat{Y}_t-V_t\theta_*)|$. Without loss of generality, we assume $\Psi_t=\{1,2,\ldots,t-1\}$ to make the expression more concise. For a fixed action $a$ in epoch $t$, we define
\begin{align*}
x_{t,a}^\top A_t^{-1}V_t^\top =&\left[\beta_1,\beta_2,\ldots,\beta_{t-1}\right],\\
\hat{\eta}_t=\widehat{Y}_{t,a}-V_t\theta_*=&\left[\hat{\eta}_t^1,\hat{\eta}_t^2,\ldots,\hat{\eta}_t^{t-1} \right].
\end{align*}

The main challenge is the bias caused by truncation such that $\E[ \hat{\eta}_t]\neq0$ and we deal with it through the inequality 
\begin{equation*}
|\hat{\eta}_t^i|\leq|\hat{\eta}_t^i-\E [\hat{\eta}_t^i]|+|\E [\hat{\eta}_t^i]|
\end{equation*}
for $i\in\{1,2,\ldots,t-1\}$, so that $\E[\hat{\eta}_t-\E [\hat{\eta}_t]]=0$ and we utilize the Bernstein's inequality \cite{Bernstein} to obtain a proper confidence interval. For the term $\E [\hat{\eta}_t^i]$, it is not a random variable and we relax it by a properly chosen truncation criterion. More details are displayed as follows. 

For any $\alpha_t'>0$, we have
\begin{equation}\label{inequality:10}
\begin{aligned}
&\mathbb{P}\left\{|x_{t,a}^\top A_t^{-1}V_t^\top (\widehat{Y}_t-V_t\theta_*)|> \alpha_t'\right\}\\
=&\mathbb{P}\left\{|\beta_1\hat{\eta}_t^1+\ldots+\beta_{t-1}\hat{\eta}_t^{t-1} |>\alpha_t'\right\}\\
\leq&\mathbb{P}\left\{\left|\beta_1\hat{\eta}_t^1+\ldots+\beta_{t-1}\hat{\eta}_t^{t-1} -\E\left[\beta_1\hat{\eta}_t^1+\ldots+\beta_{t-1}\hat{\eta}_t^{t-1} \right]\right|\right.\\
 &\left.+\left|\E\left[\beta_1\hat{\eta}_t^1+\ldots+\beta_{t-1}\hat{\eta}_t^{t-1} \right]\right|>\alpha_t'\right\}.
\end{aligned}
\end{equation}
Inspired by the technique in \citeauthor{Medina:2016} \shortcite{Medina:2016}, we have
\begin{equation}\label{ine:0325}
\begin{split}
&\left|\E\left[\beta_1\hat{\eta}_t^1+\ldots+\beta_{t-1}\hat{\eta}_t^{t-1} \right]\right|\\
=&\left|\sum_{i=1}^{t-1} \E\left[\beta_i(r_{i,a_i}\mathbbm{1}_{|\beta_ir_{i,a_i}|\leq h_{t,a}}-x_{i,a_i}^\top \theta_*)\right]\right|\\
\leq&\sum_{i=1}^{t-1} \E\left[|\beta_ir_{i,a_i}|\mathbbm{1}_{|\beta_ir_{i,a_i}|\geq h_{t,a}}\right]\\
\leq&\sum_{i=1}^{t-1} \left(\E\left[|\beta_ir_{i,a_i}|^{1+\epsilon}\right]\right)^\frac{1}{1+\epsilon}\mathbb{P}\left\{|\beta_ir_{i,a_i}|>h_{t,a}\right\}^{1-\frac{1}{1+\epsilon}}\\
\leq&\sum_{i=1}^{t-1}  |\beta_i |^{1+\epsilon}v^\frac{1}{1+\epsilon} \frac{v^{1-\frac{1}{1+\epsilon}}}{h_{t,a}^\epsilon}\\
=&\sum_{i=1}^{t-1} |\beta_i|^{1+\epsilon}\frac{v}{h_{t,a}^\epsilon}=vh_{t,a},
\end{split}
\end{equation}
where $h_{t,a}=\norm{x_{t,a}^\top A_t^{-1}V_t^\top }_{1+\epsilon}=(\sum_{i=1}^{t-1}|\beta_i|^{1+\epsilon})^{\frac{1}{1+\epsilon}}$ as defined in BTC. The second inequality is derived from H$\ddot{o}$lder inequality, the third inequality is derived from Chebyshev inequality and Assumption \ref{ass:1}. Take \eqref{ine:0325} into inequality \eqref{inequality:10}, we have
\begin{equation}\label{inequality:011}
\begin{split}
&\mathbb{P}\left\{|\beta_1\hat{\eta}_t^1+\ldots+\beta_{t-1}\hat{\eta}_t^{t-1} |>\alpha_t'\right\}\\
\leq&\mathbb{P}\left\{\left|\sum_{i=1}^{t-1}\beta_i\hat{\eta}_t^i-\E\left[\beta_i\hat{\eta}_t^i \right]\right|>\alpha_t'-vh_{t,a}\right\}\\
\leq&2\exp\left[{-\frac{(\alpha_t'-vh_{t,a})^2}{2\Var_t+\frac{2}{3}(\alpha_t'-vh_{t,a})h_{t,a}}}\right]
\end{split}
\end{equation}
through Bernstein's inequality \cite{Bernstein}, where $\Var_t=\sum_{i=1}^{t-1}\E\left[(\beta_i\hat{\eta}_t^i-\E[\beta_i\hat{\eta}_t^i])^2\right]$. The only term confuses us in inequality \eqref{inequality:011} is the variance of $\beta_i\hat{\eta}_t^i$ and we relax it as follows,
\begin{align*}
 &\sum_{i=1}^{t-1} \E\left[(\beta_i\hat{\eta}_t^i-\E[\beta_i\hat{\eta}_t^i])^2\right]\\
\leq&\sum_{i=1}^{t-1} \E\left[(\beta_ir_{i,a_i}\mathbbm{1}_{|\beta_ir_{i,a_i}|\leq h_{t,a}})^2\right]\\
=&\sum_{i=1}^{t-1} \E\left[(\beta_ir_{i,a_i}\mathbbm{1}_{|\beta_ir_{i,a_i}|\leq h_{t,a}})^{1+\epsilon}(\beta_ir_{i,a_i}\mathbbm{1}_{|\beta_ir_{i,a_i}|\leq h_{t,a}})^{1-\epsilon}\right]\\
\leq&\frac{\sum_{i=1}^{t-1}|\beta_i|^{1+\epsilon}v}{h_{t,a}^{\epsilon-1}}=h_{t,a}^2v.
\end{align*}
Thus,
\begin{align*}
 &\mathbb{P}\left\{|\beta_1\hat{\eta}_t^1+\ldots+\beta_{t-1}\hat{\eta}_t^{t-1}|>\alpha_t'\right\}\\
\leq&2\exp\left[{-\frac{(\alpha_t'-vh_{t,a})^2}{2h_{t,a}^2v+\frac{2}{3}(\alpha_t'-vh_{t,a})h_{t,a}}}\right].
\end{align*}
If we replace $\alpha_t'$ with $ \alpha_t''h_{t,a} $, where $\alpha_t''=\alpha_t'/h_{t,a}$, we have
\begin{align*} 
&\mathbb{P}\left\{|\beta_1\hat{\eta}_t^1+\ldots+\beta_{t-1}\hat{\eta}_t^{t-1}|>\alpha_t'' h_{t,a}\right\}\\
\leq&2\exp\left[-\frac{(\alpha_t''-v)^2}{2v+\frac{2}{3}(\alpha_t''-v)}\right].
\end{align*}
We display a fact, 
\begin{equation}\label{fact}
\begin{split}
c_{t,a}=&\left(x_{t,a}^\top A_t^{-1}x_{t,a}\right)^\frac{1}{2}\\
=&\left(x_{t,a}^\top A_t^{-1}(I_d+V_t^\top V_t)A_t^{-1}x_{t,a}\right)^\frac{1}{2}\\
\geq&\left(x_{t,a}^\top A_t^{-1}V_t^\top V_tA_t^{-1}x_{t,a}\right)^\frac{1}{2}\\
=&\norm{x_{t,a}^\top A_t^{-1}V_t^\top }_2\\
\geq&t^{\frac{\epsilon-1}{2(1+\epsilon)}}\norm{ x_{t,a}^\top A_t^{-1}V_t^\top }_{1+\epsilon},
\end{split}
\end{equation}
where the last inequality is due to H$\ddot{o}$lder inequality.\footnote{H$\ddot{o}$lder inequality: $\sum_{k=1}^{n}|x_ky_k|\leq\left(\sum_{k=1}^{n}|x_k|^p\right)^{1/p}\times\left(\sum_{k=1}^{n}|y_k|^q\right)^{1/q}$ for $p>0,q>0,1/p+1/q=1$. Here we take $x_k=1,y_k=\beta_k^{1+\epsilon},p=\frac{2}{1-\epsilon},q=\frac{2}{1+\epsilon}$ and $n=t-1$. } Let $2\exp\left[-\frac{(\alpha_t''-v)^2}{2v+\frac{2}{3}(\alpha_t''-v)}\right]=\frac{\delta}{TK}$, then $\alpha_t''=\frac{1}{3}\ln\frac{2TK}{\delta}+\sqrt{\frac{1}{9}\left(\ln\frac{2TK}{\delta}\right)^2+2\ln\frac{2TK}{\delta}v}+v$. With the fact $h_{t,a}\leq t^{\frac{1-\epsilon}{2(1+\epsilon)}}c_{t,a}$ proved in inequality \eqref{fact}, we get
$$\mathbb{P}\left\{|\beta_1\hat{\eta}_t^1+\ldots+\beta_{t-1}\hat{\eta}_t^{t-1}|>\alpha_t'' t^{\frac{1-\epsilon}{2(1+\epsilon)}}c_{t,a}\right\}\leq\frac{\delta}{TK}.$$
Let $\alpha_t=\left(\frac{2}{3}\ln\frac{2TK}{\delta}+\sqrt{2\ln\frac{2TK}{\delta}v}+v\right)t^{\frac{1-\epsilon}{2(1+\epsilon)}}$, then
$$\mathbb{P}\left\{|\beta_1\hat{\eta}_t^1+\ldots+\beta_{t-1}\hat{\eta}_t^{t-1}|>\alpha_t c_{t,a}\right\}\leq\frac{\delta}{TK}.$$
Combining inequalities \eqref{equation:1}, \eqref{lem2:right2} and union bound, we have  
\begin{equation*}
|\hat{r}_{t,a}-x_{t,a}^\top \theta_*|\leq(\alpha_t+1)c_{t,a},\ \ \forall a\in\{1,2,\ldots,K\}
\end{equation*}
with probability at least $1-\delta/T$.
$\hfill\blacksquare$

\subsection{Proof of Theorem 2}

If we set $\alpha=\left(\frac{2}{3}\ln\frac{2TK}{\delta}+\sqrt{2\ln\frac{2TK}{\delta}v}+v\right)T^{\frac{1-\epsilon}{2(1+\epsilon)}}$, following the similiar argument as Theorem 1, we have
\begin{equation*}
\begin{split}
R(T)=&\sum_{t=1}^{T}\E\left[r_{t,a_t^*}]-\E[r_{t,a_t}\right]\\
\leq&2\sqrt{T}+40(1+\alpha)\ln T\sqrt{dT}\\
=&2\sqrt{T}+40\ln T\sqrt{dT}+40\left(\frac{2}{3}\ln\frac{2TK}{\delta}+\right.\\
 &\left.\sqrt{2\ln\frac{2TK}{\delta}v}+v\right)\ln T\sqrt{d} T^{\frac{1}{1+\epsilon}}
\end{split}
\end{equation*}
with probability at least $1-\delta S$. Replace $\delta$ with $\delta/S$, we have with probability at least $1-\delta$,
\begin{equation*}
\begin{aligned}
R(T)\leq&2\sqrt{T}+40\ln T\sqrt{dT}+40\left(\frac{2}{3}\ln\frac{2TK\ln T}{\delta}+\right.\\
 &\left.\sqrt{2\ln\frac{2TK\ln T}{\delta}v}+v\right)\ln T\sqrt{d} T^{\frac{1}{1+\epsilon}}.
\end{aligned}
\end{equation*}
$\hfill\blacksquare$

\subsection{Proof of Theorem 3}
In this section, we provide the lower bound for SLB with heavy-tailed payoffs. We demonstrate a lower bound for the MAB with heavy-tailed payoffs first, where we adopt the techniques proposed by \citeauthor{Auer:2002b} \shortcite{Auer:2002b}. Then, we extend it to SLB by a proper design of the contextual feature vectors $x_{t,a}$ and coefficient vector $\theta_*$. For MAB problem, we consider the following payoff distribution,
\begin{equation}\label{MAB-design}
\begin{aligned}
r_{t,a^*}=&
\begin{cases}
1/\gamma \quad &\textnormal{with probability of }2\gamma^{1+\epsilon},\\
0\quad &\textnormal{with probability of }1-2\gamma^{1+\epsilon},
\end{cases}\\
r_{t,a}=&
\begin{cases}
1/\gamma \quad &\textnormal{with probability of }\gamma^{1+\epsilon},\\
0\quad &\textnormal{with probability of }1-\gamma^{1+\epsilon},
\end{cases}
\end{aligned}
\end{equation}
where $a^*\in\{1,\ldots,K\}$ is chosen uniformly at random and $a\neq a^*,\gamma>0$. It is easy to verify that the $1+\epsilon$ raw moments of above payoffs are bounded by 2.

\begin{lem}\label{lower bound:MAB}
For any bandits algorithm $\B$ with $T\geq K\geq 4$, an arm $a^*\in\{1,\ldots,K\}$ is chosen uniformly at random,  this arm pays $1/\gamma$ with probability $p_{a^*}=2\gamma^{1+\epsilon}$ and the rest arms pay $1/\gamma$ with probability $\gamma^{1+\epsilon}$. If we set $\gamma=(K/(T+2K))^{\frac{1}{1+\epsilon}}$, we have
\begin{equation*}
\E[p_{a^*}T-\sum_{t=1}^{T}r_{t,a_t}]\geq \frac{1}{8}T^\frac{1}{1+\epsilon}K^\frac{\epsilon}{1+\epsilon}.
\end{equation*}
\end{lem}

Above lemma provides a lower bound for heavy-tailed MAB and the detailed proof of Lemma \ref{lower bound:MAB} is provided later. 

In order to establish the lower bound for SLB under heavy-tailed setting, the payoff function of $x_{t,a}$ is designed as
\begin{equation}\label{SLB-design}
r_{t,a}=
\begin{cases}
1/\gamma \quad &\textnormal{with probability of }\gamma\cdot x_{t,a}^\top \theta_*,\\
0\quad &\textnormal{with probability of }1-\gamma\cdot x_{t,a}^\top \theta_*.
\end{cases}
\end{equation}

We notice the payoff of SLB is the same as MAB once $x_{t,a}^\top \theta_*=\gamma^\epsilon$ and $x_{t,a^*}^\top \theta_*=2\gamma^\epsilon$. Inspired by the technique of \citeauthor{Abe:2003} \shortcite{Abe:2003}, we divide all trails into $S=\lfloor(d-1)/K\rfloor$ stages with $m=\lfloor T/S \rfloor$ trails in each stage. We say the trail $t$ belong to stage $j$ if $\lfloor t/m\rfloor=j$ for $j\in\{0,1,\ldots,S-1\}$.

The contextual vector is designed as follows. For simplification, we number features from 0. Feature 0 has a value of $\sqrt{1/2}$ for all arms at all epochs. During the  $j$-th stage, the value of the $(j\times K+i)$-th feature of the $i$-th arm is also $\sqrt{1/2}$ for $i\in\{1,2,\ldots,K\}$. The rest features have a value of 0. 

For SLB problem with above contextual information, any algorithm $\A$ with $T$ trails can be regarded as a sequence of algorithms $\B_1,\B_2,\ldots,\B_S$. The algorithm $\A$ starts up by a random input at the begining of $j$-th stage, then the strategy adopted by algorithm $\A$ during this stage is equivalent to the algorithm $\B_j$ for solving bandit problem.


Now, we set the coefficients of the linear bandits $\theta_*=[\theta_0,\theta_1,\ldots,\theta_{d-1}]$ as follows. Let $\theta_0=\sqrt{2}\gamma^\epsilon$ and for the $j$-th stage, we choose $i_j$ uniformly at random from $\{1,2,\ldots,K\}$, then the value of $\theta_{j\times K+i_j}$ is $\sqrt{2}\gamma^\epsilon$ and the rest elements in $\{\theta_{j\times K+1},\ldots,\theta_{(j+1)\times K}\}$ are $0$. With feature vectors and coefficients vector as above, the expected payoff for $i_j$-th arm during the $j$-th stage is $2\gamma^\epsilon$, the rest arms have the expected payoff $\gamma^\epsilon$. With the lower bound for MAB in Lemma \ref{lower bound:MAB}, we have
\begin{equation*}
\E[R(T)]\geq\frac{1}{8}S\left(\left\lfloor\frac{T}{S}\right\rfloor\right)^{\frac{1}{1+\epsilon}}K^\frac{\epsilon}{1+\epsilon}\geq\frac{1}{32}\left(d-1\right)^{\frac{\epsilon}{1+\epsilon}}T^{\frac{1}{1+\epsilon}},
\end{equation*} 
where the expectation is with respect to the random choice of $\theta_*$ and the randomness of the learning process. Thus,
\begin{equation*}
\E[R(T)]\geq O\left(d^{\frac{\epsilon}{1+\epsilon}}T^{\frac{1}{1+\epsilon}}\right).
\end{equation*} 
The proof is finished.
$\hfill\blacksquare$

Through the design of contextual vector, we conjecture that the complexity of MAB and SLB problems are depending on the number of arms $K$ and feature dimenson $d$, respectively. This is a reasonable conjecture since the number of inherent variables in MAB and SLB are $K$ and $d$ respectively. To close the gap between upper bound and lower bound, it might be workable to design an algorithm for MAB with upper regret bound $O(K^{\frac{\epsilon}{1+\epsilon}}T^{\frac{1}{1+\epsilon}})$ or figure out a situation with lower regret bound $O(K^{\frac{1}{2}}T^{\frac{1}{1+\epsilon}})$. We are working hard to close this gap.

\subsection{Proof of Lemma 7}

First of all, we define some new notations. $\r^t =\langle r_{1,a_1},\ldots,r_{t,a_t}\rangle$ is the payoff sequence and $\r^T $ is simplified as $\r$. $\P\{\cdot\}$ is the probability with respect to random choice of payoffs and $\P_a\{\cdot\}=\P\{\cdot|I=a\}$ is the probability conditioned on $a$ being the good arm. $\P_{unif}\{\cdot\}$ is the probability with respect to the payoffs that all arms associated with the same distribution. $\E[\cdot],\ \E_a[\cdot],\ \E_{unif}[\cdot]$ are the expection with respect to above probability. $G_\B=\sum_{t=1}^T  r_{t,a_t}$ is the return of the alogorithm $\B$ and $G_{max}=\max_a\sum_{t=1}^{T}r_{t,a}$ is the return of the best action. We propose a useful lemma as follows.

\begin{lem}\label{lem:lowerI}
Let $f:\mapsto[0,\ M]$ be any function defined on payoff sequence $\r$. Then, for any arm $a$,
\[
\E_a[f(\r)]\leq\E_{unif}[f(\r)]+\frac{M}{2}\sqrt{\ln2\times\E_{unif}[N_a]\frac{\gamma^{1+\epsilon}}{1-2\gamma^{1+\epsilon}}},
\]
where $N_a$ is the number of times to choose arm $a$ in the sequence $\r$.
\end{lem}

\noindent\textbf{Proof of Lemma \ref{lem:lowerI}} We follow the standard methods which can be found, for instance, in \cite{Auer:2002b}. We define the variational distance for any distribution $\P$ and $\Q$ as
\[
\norm{\P-\Q}_1=\sum_{\r\in\{0,\gamma\}^T }|\P\{\r\}-\Q\{\r\}|
\]
and let
\[
KL(\P\parallel\Q)=\sum_{\r\in\{0,\gamma\}^T }\P\{\r\}\log_2\left(\frac{\P\{\r\}}{\Q\{\r\}}\right)
\]
be the Kullback-Liebler divergence between the two distributions. Meanwhile, we define the KL-divergence for the payoff at epoch $t$ conditioned on the previous payoffs as
\begin{align*}
&KL(\P\{r_{t,a_t}|\r^{t-1}\}\parallel\Q\{r_{t,a_t}|\r^{t-1}\})\\
=&\sum_{\r\in\{0,\gamma\}^T }\P\{r_{t,a_t}|\r^{t-1}\}\log_2\left(\frac{\P\{r_{t,a_t}|\r^{t-1}\}}{\Q\{r_{t,a_t}|\r^{t-1}\}}\right).
\end{align*} 
For the Bernoulli random variables with parameters $p$ and $q$, we use 
\[
KL(p\parallel q)=p\log_2\left(\frac{p}{q}\right)+(1-p)\log_2(\frac{1-p}{1-q})
\]
to denote the KL-divergence related to these two distributions. We have
\begin{equation}\label{main}
\begin{split}
&\E_a[f(\r)]-\E_{unif}[f(\r)]\\
=&\sum_{\r\in\{0,\gamma\}^T }f(\r)(\P_a\{\r\}-\P_{unif}\{\r\})\\
\leq&\sum_{\r:\P_a\{\r\}\geq\P_{unif}\{\r\}}f(\r)(\P_a\{\r\}-P_{unif}\{\r\})\\
\leq&M\sum_{\r:\P_a\{\r\}\geq\P_{unif}\{\r\}}\P_a\{\r\}-P_{unif}\{\r\}\\
=&\frac{M}{2}\norm{P_a-P_{unif}}_1.
\end{split}
\end{equation} 
Lemma 12.6.1 in \citeauthor{Cover:2006} \shortcite{Cover:2006} states that
\begin{equation}\label{KL-inequality1}
\norm{P_a-P_{unif}}_1^2\leq(2\ln2)KL(\P_{unif}\parallel\P_a).
\end{equation}
Based on the chain rule for KL-divergence [\citeauthor{Cover:2006} \shortcite{Cover:2006} Theorem 2.5.3], we have 
\begin{equation}\label{KL-inequality2}
\begin{split}
 &KL(\P_{unif}\parallel\P_a)\\
=&\sum_{t=1}^{T}KL(\P_{unif}\{r_{t,a_t}|\r^{t-1}\}\parallel\P_a\{r_{t,a_t}|\r^{t-1}\})\\
=&\sum_{t=1}^{T}\P_{unif}\{a_t\neq a\}KL(\gamma^{1+\epsilon}\parallel\gamma^{1+\epsilon})\\
 &+\P_{unif}\{a_t= a\}KL(\gamma^{1+\epsilon}\parallel2\gamma^{1+\epsilon})\\
=&\sum_{t=1}^{T}\P_{unif}\{a_t= a\}KL(\gamma^{1+\epsilon}\parallel2\gamma^{1+\epsilon})\\
\leq&\E_{unif}[N_a]\frac{\gamma^{1+\epsilon}}{2(1-2\gamma^{1+\epsilon})},
\end{split}
\end{equation}
where the last inequality is due to $KL(p\parallel q)\leq\frac{(p-q)^2}{q(1-q)}$. 

We give a simple explanation about the second equality in formula \eqref{KL-inequality2}. The conditional distribution $\P_{unif}\{r_{t,a_t}|\r^{t-1}\}$ is a Bernoulli distribution with parameter $\gamma^{1+\epsilon}$. The conditional distribution $\P_a\{r_{t,a_t}|\r^{t-1}\}$ depends on the good arm $a$. The arm $a_t$ is fixed by the algorithm given the historical payoffs $\r^{t-1}$. If the arm is not the good arm $a$, the Bernoulli parameter is $\gamma^{1+\epsilon}$. Otherwise, the $r_{t,a_t}$ is $1/\gamma$ with probality $2\gamma^{1+\epsilon}$. Thus, the proof of Lemma \ref{lem:lowerI} is finished by combining \eqref{main}, \eqref{KL-inequality1} and \eqref{KL-inequality2}.
$\hfill\blacksquare$

\noindent\textbf{Proof of Lemma \ref{lower bound:MAB}} If arm $a$ is chosen to be the best arm, then the expected payoff at time $t$ is 
\begin{equation*}
\begin{split}
\E_a[r_{t,a_t}]=&2\gamma^{\epsilon}\P_a\{a_t=a\}+\gamma^{\epsilon}\P_a\{a_t\neq a\}\\
=&2\gamma^{\epsilon}\P_a\{a_t=a\}+\gamma^{\epsilon}(1-\P_a\{a_t=a\})\\
=&\gamma^{\epsilon}+\gamma^{\epsilon}\P_a\{a_t=a\}.
\end{split}
\end{equation*}
Thus, the expected return of algorithm $\A$ is
\begin{equation}\label{expected-pay}
\E_a[G_\B]=\sum_{t=1}^T \E_a[r_{t,a_t}]=T\gamma^{\epsilon}+\gamma^{\epsilon}\E_a[N_a].
\end{equation}
We apply Lemma \ref{lem:lowerI} to the $N_a$, which is a function the payoff sequence $\r$. With $N_a\in[0,T]$, we have
\begin{equation*}
\E_a[N_a]\leq\E_{unif}[N_a]+\frac{T}{2}\sqrt{\ln2\times\E_{unif}[N_a]\frac{\gamma^{1+\epsilon}}{1-2\gamma^{1+\epsilon}}}.
\end{equation*}
Summing over all arms gives
\begin{equation*}
\begin{split}
&\sum_{a=1}^{K}\E_a[N_a]\\
\leq&\sum_{a=1}^{K}\left(\E_{unif}[N_a]+\frac{T}{2}\sqrt{\ln2\times\E_{unif}[N_a]\frac{\gamma^{1+\epsilon}}{1-2\gamma^{1+\epsilon}}}\right)\\
\leq&T+\frac{\sqrt{\ln2}\cdot T}{2}\sqrt{KT\frac{\gamma^{1+\epsilon}}{1-2\gamma^{1+\epsilon}}}
\end{split}
\end{equation*}
 by the fact $\sum_{a=1}^{K}\E_{unif}[N_a]=T$ and $\sum_{a=1}^{K}\sqrt{\E_{unif}[N_a]}\leq\sqrt{KT}$. If we combine above inequality with inequality \eqref{expected-pay}, then we have
\begin{equation*}
\begin{aligned}
\E[G_\B]=&\frac{1}{K}\sum_{a=1}^{K}\E_a[G_\B]\\
\leq& T\gamma^{\epsilon}+\gamma^{\epsilon}\left(\frac{T}{K}+\frac{\sqrt{\ln2}\cdot T}{2}\sqrt{\frac{T}{K}\frac{\gamma^{1+\epsilon}}{1-2\gamma^{1+\epsilon}}}\right).
\end{aligned}
\end{equation*}
The expected payoff of the best arm is 
\begin{equation*}
\E[G_{max}]\geq 2\gamma^\epsilon T.
\end{equation*}
Therefore, we have 
\begin{equation*}
\begin{aligned}
 &\E[G_{max}]-\E[G_\B]\\
\geq& T\gamma^{\epsilon}\left(1-\frac{1}{K}-\frac{\sqrt{\ln2}}{2}\sqrt{\frac{T}{K}\frac{\gamma^{1+\epsilon}}{1-2\gamma^{1+\epsilon}}}\right).
\end{aligned}
\end{equation*}
Replace $\gamma$ with $(K/(T+2K))^{\frac{1}{1+\epsilon}}$ and if $T\geq K\geq 4$, lower bound is established as
\begin{equation*}
\E[G_{max}]-\E[G_\B]\geq \frac{1}{8}T^\frac{1}{1+\epsilon}K^\frac{\epsilon}{1+\epsilon}.
\end{equation*}
The proof of Lemma \ref{lower bound:MAB} is finished.
$\hfill\blacksquare$

\begin{figure*}[tb]
\centering 
\subfigure[Student's $t$-Noise]{
\includegraphics[width=0.45\textwidth,height=0.25\textwidth]{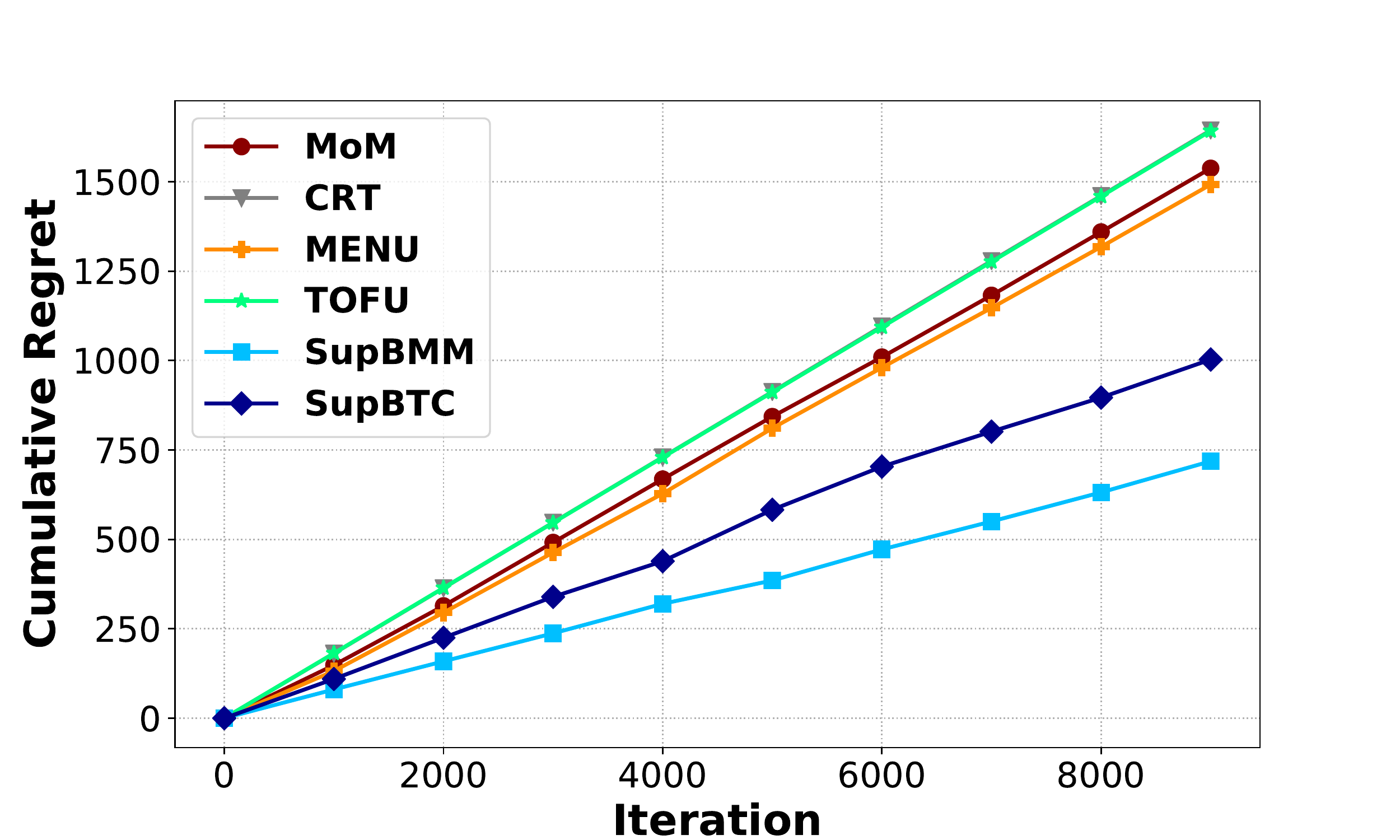}}
\subfigure[Pareto Noise]{
\includegraphics[width=0.45\textwidth,height=0.25\textwidth]{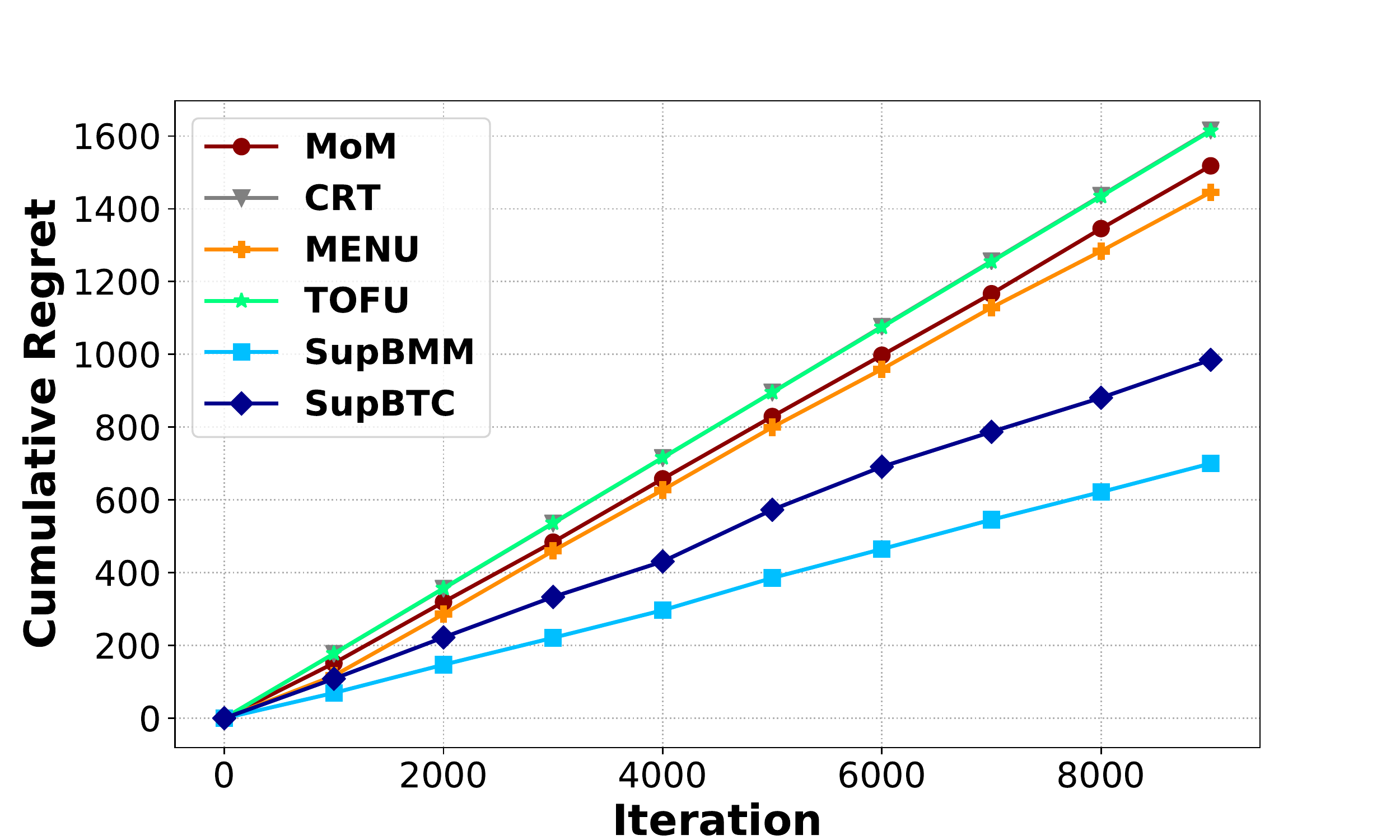}}

\caption{Comparison of our algorithms versus the MoM, CRT, MENU and TOFU.}
\label{Fig-2016}
\end{figure*}

\section{Experiments}
In this section, we conduct experiments to evaluate the proposed algorithms. All algorithms' parameters are set to $\epsilon=1$ and $\delta=0.01$. We adopt MoM and CRT of \citeauthor{Medina:2016} \shortcite{Medina:2016}, MENU and TOFU of \citeauthor{Han:2018} \shortcite{Han:2018} as baselines for comparison. 

Let the feature dimension $d = 10$, the number of arms $K = 20$ and $\theta_*=\ones/\sqrt{d}\in\R^d$, where $\ones$ is an all-$1$ vector so that $\norm{\theta_*}=1$. Each element of the vector $x_{t,a}$ is sampled from the uniform distribution of $[0,1]$, and then the vector is normalized to a unit vector ($\norm{x_{t,a}}=1$). According to the linear bandit model, the observed payoff is 
$$r_{t,a}=x_{t,a}^\top\theta_*+\eta_t$$
where $\eta_t$ is generated from the following two noises.

\begin{enumerate}[(i)]
\item Student's $t$-Noise: The probability density function of this noise is $\eta_t\sim\frac{\Gamma(2)}{\sqrt{3\pi}\Gamma(1.5)}\left(1+\frac{x^2}{3}\right)^{-2}$ for $x\in\R$ and $\Gamma(\cdot)$ is the Gamma function. Thus, the bounds of the second central moment and second raw moment of payoff are $3$ and $4$, respectively. 

\item Pareto Noise: The probability density function of this noise is $\eta_t\sim\frac{sx_m^s}{x^{s+1}}\mathbbm{1}_{x\geq x_m}$ for $x\in\R$ and we set the shape $s=3$ and the scale $x_m=0.01$. The bounds of the second central moment and second raw moment of payoff are $1$ and $2$. 
\end{enumerate}
The main difference between the above two heavy-tailed noises is that Student's $t$-distribution is symmetric while Pareto distribution is not.

We run 10 independent repetitions for each algorithm and display the average cumulative regret with time evolution. Fig.~\ref{Fig-2016}(a) compares our algorithms against algorithms of \citeauthor{Medina:2016} \shortcite{Medina:2016} and \citeauthor{Han:2018} \shortcite{Han:2018} under Student's $t$-noises. Fig.~\ref{Fig-2016}(b) presents the cumulative regrets under Pareto noises. Our algorithms outperform MoM, CRT, MENU and TOFU with the interference of symmetric or asymmetric noises, which verifies the effectiveness of our algorithms on the heavy-tailed bandit problem. SupBMM achieves the smallest regret which is expected,since compared to SupBTC it has a more favorable logarithmic factor in regret bound.

\section{Conclusion and Future Work}
In this paper, we develop two novel algorithms to settle the heavy-tailed issue in linear contextual bandit with finite arms. Our algorithms only require the existence of bounded $1+\epsilon$ moment of payoffs, and achieve $\widetilde{O}(d^{\frac{1}{2}}T^{\frac{1}{1+\epsilon}})$ regret bound which is tighter than that of \citeauthor{Han:2018} \shortcite{Han:2018} by an $O(\sqrt{d})$ factor for finite action sets. Furthermore, we provide a lower bound on the order of $\Omega(d^\frac{\epsilon}{1+\epsilon}T^\frac{1}{1+\epsilon})$. Finally, our proposed algorithms have been evaluated based on numerical experiments and the empirical results demonstrate the effectiveness in addressing heavy-tailed problem.

In the future, we will investigate more on closing the gap between upper bound and lower bound with respect to the dimension $d$.

\bibliographystyle{named}
\bibliography{ref}

\end{document}